# Practical Applications of Advanced Cloud Services and Generative AI Systems in Medical Image Analysis

**Jingyu Xu 1***

Computer Information Technology,Northern Arizona University,Flagstaff, AZ,USA

* Corresponding author: jyxu01@outlook.com

**Binbin Wu 2**

Heating Ventilation and Air Conditioning Engineering,   Tsinghua University, Beijing China

wubinbin.1@gmail.com

**Jiaxin Huang 3**

Information Studies,Trine University,Phoenix USA

jiaxinhuang1013@gmail.com

**Yulu Gong 4**

Computer & Information Technology,Northern Arizona University,Flagstaff, AZ, USA

yg486@nau.edu

**Yifan Zhang 5**

Executive Master of Business Administration, Amazon Connect Technology Services (Beijing) Co., Ltd.

 Xi'an, Shaanxi, China

yifan.ibm@gmail.com

**Bo Liu 6**

Software Engineering,Zhejiang University,HangZhou,China

lubyliu45@gmail.com

**Abstract**

The medical field is one of the important fields in the application of artificial intelligence technology. With the explosive growth and diversification of medical data, as well as the continuous improvement of medical needs and challenges, artificial intelligence technology is playing an increasingly important role in the medical field. Artificial intelligence technologies represented by computer vision, natural language processing, and machine learning have been widely penetrated into diverse scenarios such as medical imaging, health management, medical information, and drug research and development, and have become an important driving force for improving the level and quality of medical services.The article explores the transformative potential of generative AI in medical imaging, emphasizing its ability to generate synthetic data, enhance images, aid in anomaly detection, and facilitate image-to-image translation. Despite challenges like model complexity, the applications of generative models in healthcare, including Med-PaLM 2 technology, show promising results. By addressing limitations in dataset size and diversity, these models contribute to more accurate diagnoses and improved patient outcomes. However, ethical considerations and collaboration among stakeholders are essential for responsible implementation. Through experiments leveraging GANs to augment brain tumor MRI datasets, the study demonstrates how generative AI can enhance image quality and diversity, ultimately advancing medical diagnostics and patient care.

## INTRODUCTION

Generative AI is a series of AI techniques and models designed to learn the underlying patterns and structures of data sets and generate new data points that may belong to the original data set. Although the data distribution of high-dimensional imaging data is very complex, today we have a variety of powerful deep generation models, such as diffusion models, autoregressive [1]Transformers, generative adversarial networks (GANs), and variational autoencoders (VAEs), that have proven effective at learning and representing these intricate data distributions.

Generative AI, exemplified by ChatGPT, has demonstrated a high ability to understand the semantics and context of natural language and generate coherent, meaningful responses, but it is uncertain how well it will perform in the medical field, which requires a high level of expertise and complex thinking. Combined with the current status of China's primary medical and health service ability, the author discusses the potential application of generative artificial intelligence technology represented by ChatGPT in the improvement of primary medical and health service ability, including clinical assistant diagnosis and treatment, writing electronic medical records, remote management of chronic disease patients and patient education. However, its own limitations such as inability to guarantee accuracy, lack of doctor-patient interaction, language barriers, as well as data security, patient privacy and ethical issues limit its application. For the application of ChatGPT in primary health care services, the whole society needs to discuss and analyze the potential risks, and formulate corresponding policies and regulations to ensure the introduction and application of ChatGPT in a prudent and responsible manner, so as to achieve the goal of enhancing the ability of primary health care services.

The ability to generate synthetic data has been one of the main goals of generating models in medical imaging, as it offers a viable solution for sharing data while protecting patient privacy. Data providers can train models on their own data and share results in a way that preserves privacy, rather than directly sharing patient data. Recent medical imaging studies have demonstrated the promising ability to generate high-quality medical images. In addition, machine learning models trained for downstream tasks using datasets that include synthetic data have achieved similar or even superior performance to models trained using only real data

In addition to generating synthetic data, generative models have a wide range of other applications in medical imaging. One of these is anomaly detection, generating models that can be used to identify anomalies in medical images. This task is particularly useful for diagnosing diseases and detecting potential health risks. Another application is image-to-image translation (style transfer), where generative models can be trained to convert images from one mode to another. For example, these models can be used to convert a CT [2-5]scan to an MRI scan, a label map to an MRI image, or an MRI scan to an anomaly map. In addition, generative models can be used for image enhancement, where they can learn to improve the quality of medical images without losing important clinical information. This includes image denoising, image artifact removal, image resolution enhancement, etc. Finally, the generated model can also be used for image reconstruction, including MRI and CT reconstruction, which can reconstruct high-quality images from under-sampled or noisy data. This is of great significance for medical imaging and can speed up human imaging. Overall, the potential applications of generative models in medical imaging are very broad, and their use is likely to continue to grow.

However, this area is accompanied by challenges, including the use of various quality assessment indicators and the complexity of the models. These issues may hinder the progress and implementation of the model.

# 1 RELATED WORK

## 1.1 Generative AI

Generated type artificial intelligence (generativeartificialintelligence, referred to as "emergent AI), also known as" artificial intelligence to generate content, "the goal is to learn a great deal of training data to understand the probability distribution of the data, then use this understanding to generate new, similar to the content of the training data, such as images, text, Audio, etc. Some common generative[6] AI models include autoregressive models, generative adversarial networks, variational autoencoders, etc. Among them, the common autoregressive models are recurrent neural networks and converters. Generative artificial intelligence tries to act as a scaffolding or midwife to activate people's subjective potential, realize mutual enlightenment of human intelligence and wit, promote the symbiotic development of different educational forms, digital technologies and educational resources, and incubate new opportunities for the systematic transformation of the digital transformation of ideological and political education.

Generative AI is a powerful technology that can automatically generate various forms of information, including text, code, images, audio and video, by learning and mimicking the way humans create content. The application scenarios for generative AI are very broad, and it has shown great potential and value in many areas. This article takes stock of the application scenarios of generative AI in different fields to demonstrate its important role in driving technological innovation and social development.

Compared with traditional "analyticalAI", generative AI has unique advantages. The most important feature is its ability to create new content, not just the identification or classification of existing data. Generative AI systems describe the distribution of training data by building probabilistic models. This can be a model based on statistical methods, or it can be a deep learning model based on neural networks. The goal of the model is to capture the internal structure and regularity of the training data.[7-9] Once a generative AI model learns the distribution of the training data, it can be used to generate new data samples that are somewhat similar to the training data, but with certain differences. This generative ability makes generative AI very useful for things like natural language processing (generating text), computer vision (generating images and video), audio processing (generating music and speech), drug discovery, artistic creation, and more. They are also used for data enhancement, sample generation, simulation and other tasks. This will have a profound impact on the field of artificial intelligence and accelerate the arrival of the era of strong artificial intelligence. So, what different dynamics and risks will the rise of generative AI bring to the field of stomatology? In view of its potential, it is necessary for us to think deeply about its development prospects in the field of stomatology.

## 1.2 Traditional medical imaging challenges

In 2018, OpenAI introduced the first generation of generative pre-trained transformer (GPT), bringing natural language processing into the "pre-training" era. With the rapid development of deep learning and natural language processing technology, the first generation of GPT has gradually evolved into ChatGPT-4, which can understand natural language input and generate fluent natural language output. ChatGPT uses a deep learning algorithm based on Transformer architecture, which is based on technologies such as self-attention mechanism, residual linkage and layer normalisation to automatically learn semantic and grammatical rules in text.In light of recent developments, this review article provides an overview of existing industry and research efforts in the application of generative AI models in medicine and healthcare. By highlighting the enormous potential, benefits, challenges, and ethical considerations of AI, this study aims to foster an ongoing dialogue about responsibly using AI's transformative power to improve medical practice and patient health. As we explore the impact of generative models on healthcare and medicine, adherence to ethical principles, patient-centered healthcare, and collaboration among AI developers, medical staff, and policy makers will be critical to exploring evolving healthcare AI.

For now, most MRI and PET-CT are currently operating at full capacity, and patients often need at least three days or several weeks to book an MRI, if it is in Beijing, Shanghai and other places, the waiting time will be further extended. The core reasons are mainly reflected in two aspects: one is the huge gap in the number of diagnostic doctors and technicians; Second, the detection capacity of image detection equipment is insufficient, and a scan takes 30 minutes to 1 hour. In addition, in the face of the imaging needs of some specific examinations, it is necessary to inject contrast agents for patients, the use of contrast agents to a certain extent, resulting in the deposition of contrast agents in the patient's body, causing radiation damage, so it is also necessary to consider how to reduce the use of contrast agents, on the other hand, the reduction of dose can also reduce the cost of inspection.

Finally, for most of the current hospitals, in order to solve the problem of patient queuing mentioned earlier, it is necessary to purchase imaging equipment with higher scanning speed and quality, but a high-end imaging equipment costs tens of millions, or upgrading existing equipment also needs to invest millions, but this is only the cost of money, and the time cost is high.[10] It is reported that the cycle of upgrading or purchasing imaging equipment is about six months to one year, and in the process, it will cause the limited service capacity of the hospital, and cause some patients to choose other hospitals because they do not want to continue to wait. In addition, additional equipment requires venues and supporting medical personnel, which requires additional costs.

Therefore, taking into account the downstream problem, when the image quality is low in the actual situation, it will affect the quantitative diagnosis effect based on the image in the later stage, and then affect the final treatment effect. In this regard, Dr. Xiang Lei said, "When downstream service providers do quantitative analysis based on image data, if the quality of the medical image itself is not high, then the quantitative analysis obtained will be greatly affected."

## 1.3 Generative artificial intelligence and healthcare

The application of generative intelligence in the field of medical imaging is developing rapidly. At present, generative intelligence has been applied to many aspects of medical imaging, including image generation, image enhancement, pathological image analysis, image recognition and auxiliary diagnosis. These applications use a combination of deep learning and generative models to provide healthcare professionals with more accurate and efficient image analysis and diagnosis. The application of generative intelligence in the field of medical imaging has achieved some remarkable results. For example, generative models such as generative Adversarial networks (GANs) are used to synthesize photorealistic medical image data to increase the diversity and number of data sets, thereby improving the generalization ability and performance of the model. In addition, the generative model can also be used to enhance medical images, such as denoising, enhancing contrast and sharpness, so as to improve image quality and visualization effect. In addition, generative intelligence is also being applied to the segmentation and feature extraction of medical images to help doctors more accurately identify and locate diseased areas.

In this context, Med-PaLM 2 technology introduces advanced generative intelligent algorithms, combined with expertise and technology in the field of medical imaging, aiming to further advance the analysis, diagnosis and treatment of medical imaging. Med-PaLM 2 [11]May include features such as more efficient medical image synthesis, more accurate lesion detection and segmentation, and more reliable pathological image analysis to provide healthcare workers with a more powerful and comprehensive image analysis tool.

Google is currently developing the ability of Med-PaLM 2 to analyze X-ray and mammogram images, allowing AI to add to the recent boom in generative AI by analyzing output reports and responding to follow-up questions. Google has not yet demonstrated LLM applications in healthcare Settings, and it is not yet open for use. Med-PaLM 2 seems to be a promising application of AI in healthcare. Existing models either analyze images or answer questions. By combining these two capabilities in the same system, Google will allow physicians to challenge the conclusions made by LLMS.

Relevant research reports that Med-PaLM 2 is one of the latest medical image detection technologies in the field of generative intelligence, and the biggest difference from the existing traditional AI image analysis system, which provides conclusions after analyzing the image, but does not explain the reason. In theory, physicians could talk to Google's LLM like a colleague, discussing images and initial assessments and turning the discussion into an ongoing conversation.

## 2 EXPERIMENT AND METHODOLOGY

Generative AI, encompassing advanced models like Generative Adversarial Networks (GANs), Variational Autoencoders (VAEs), and autoregressive Transformers, offers a unique set of advantages in the context of medical image analysis. Medical imaging datasets are often limited in size and diversity, posing challenges for training accurate and robust machine learning models. Generative AI addresses this issue by synthesizing realistic and diverse medical images, thereby augmenting the available dataset. [12]This augmentation not only increases the quantity of data but also enhances its diversity, enabling more comprehensive training of machine learning models. Furthermore, generative models can be used for anomaly detection, a critical task in medical imaging, where abnormalities are often subtle and challenging to identify. By learning the underlying patterns of normal and abnormal anatomy, generative AI can assist radiologists and clinicians in detecting and localizing anomalies with greater accuracy and efficiency.

### 2.1 Experimental design

In this study, we harness the power of generative AI, specifically GANs, to address the challenges associated with limited and heterogeneous brain tumor MRI datasets.By generating synthetic brain tumor MRI images, we aim to augment the existing dataset and improve the performance of tumor detection and classification algorithms.Through this approach, we seek to leverage the strengths of generative AI to enhance the accuracy, efficiency, and reliability of brain tumor diagnosis, ultimately contributing to improved patient outcomes and clinical decision-making in neuro-oncology.

## 2.2 Generative Modelling

At the current population level of the country (1.417 billion), this means only 0.0035 percent are diagnosed with Brain Tumor!

Let's assume that all MRI scans produce 100% accurate results. This would mean that for every 10,000 MRI scans, we only get 35 samples showing Brain Tumor versus many more that don't

This, combined with other problems in accessing Medical data, would lead to Machine Learning problems such as Class Imbalance and Bias

Source: https://health.economictimes.indiatimes.com/news/diagnostics/brain-tumors-death-on-diagnosis/88090467

Generative adversarial networks are implicit likelihood models that generate data samples from the statistical distribution of the data. They're used to copy variations within the dataset. They use a combination of two networks: generator and discriminator.

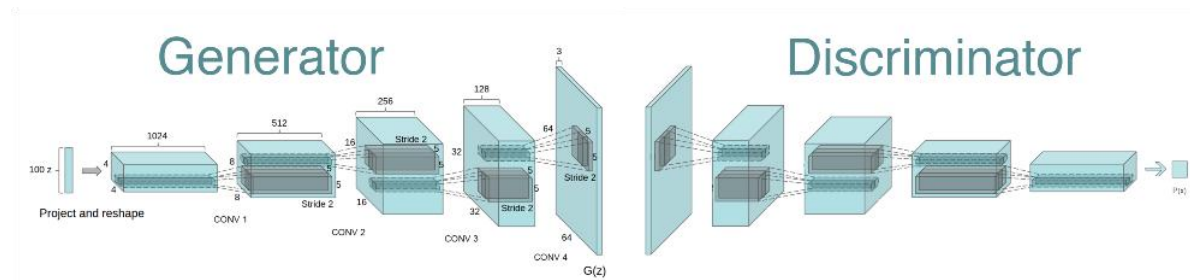

**figure1.** Generative adversarial networks are implicit likelihood models

## 2.3 loss function or the value function

$$\nabla_{\theta_d} \frac{1}{m} \sum_{i=1}^{m} [\log D\left(x^{(i)}\right) + \log \left(1 - D\left(G\left(Z^{(i)}\right)\right)\right)] \quad (1)$$

It measures the distance between the distribution of the data generated and the distribution of the real data.Both the generator and the discriminator have their own loss functions.The generator tries to minimize the loss function while the discriminator tries to maximize.

## 2.4 Image data preprocessing

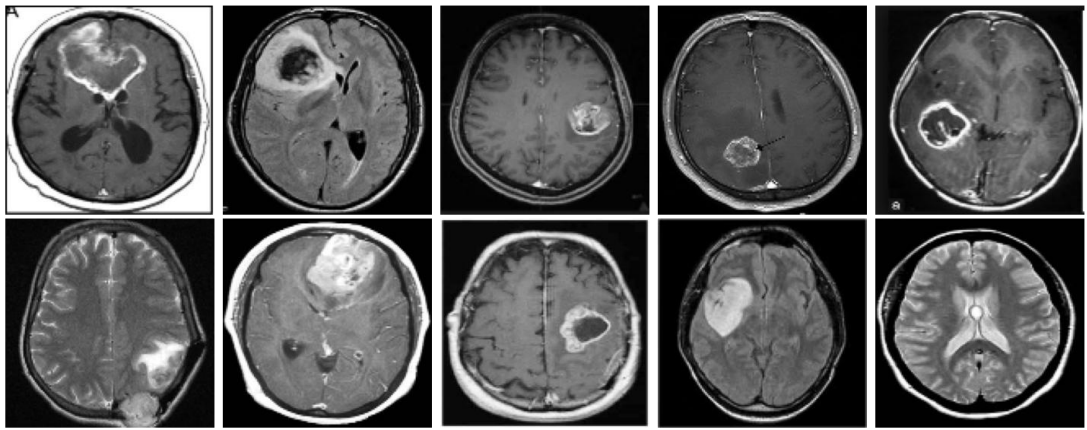

**Figure 2.** Raw image data processing results

This section loads a dataset of MRI images containing brain tumors through the load_images function and converts them to grayscale images, then adjusts them to the specified size (128x128). Then, the image is normalized, the pixel values are scaled to the [-1, 1] range, and the image data is realigned to fit the model input. In addition, the Matplotlib library was used to draw 20 real brain tumor MRI image samples from the data set and display them in the form of gray color maps.

## 2.5 Model building and training

Step 1: Build a Generator model: Define a generator model that takes a random noise vector as input and converts the noise vector into an image through multiple Conv2DTranspose layers. The final output of the generator is a composite image that resembles the real image.

Step 2: Build a Discriminator model: A discriminator model is defined that distinguishes the synthetic image generated by the generator from the real image. The discriminator classifies the input image as real or false through multiple convolution layers (Conv2D) and fully connected layers (Dense).

Combine generator and discriminator into GAN model:[13] Combine generator and discriminator into one overall generative adversarial network (GAN) model. In a GAN model, the output of the generator is fed into a discriminator that evaluates the authenticity of the generated image.

Step 3: Define a generator sample visualization function: Define a function, sample_images, to generate and visualize the image samples generated by the generator model. The function takes a random noise vector as input and uses the generator model to generate a corresponding number of composite images, which are displayed in a specified format.

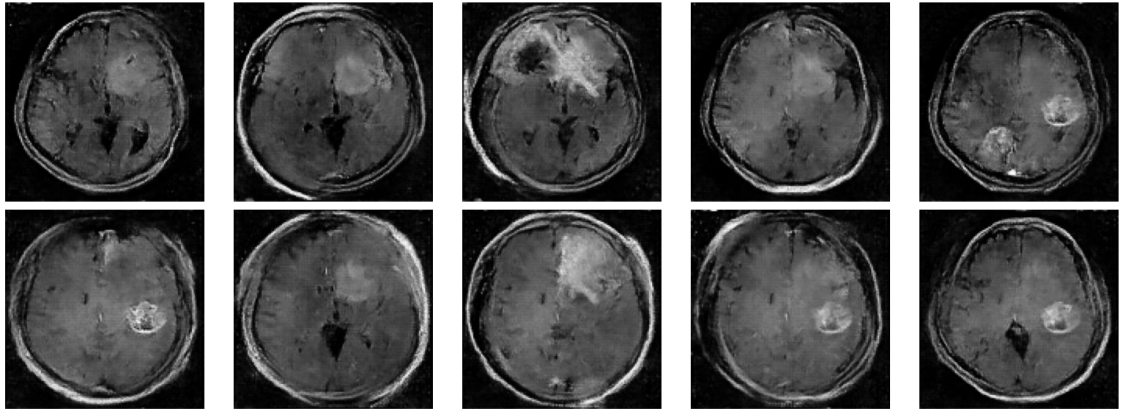

**figure 3.** Training image results

The following can be learned from the training image results:

1.Generator Loss: With the progress of training, the loss of the generator gradually decreases, indicating that the generator gradually learns to generate more realistic images.

2.Discriminator Loss: The discriminator loss may fluctuate during training, but the overall trend is gradually decreasing, indicating that the discriminator gradually learns to distinguish between real and generated images.

3. Training progress: As the training progresses, the steps within each epoch (batch) are gradually completed, while the loss of generators and discriminators is also changing. To a certain extent, the training of generator and discriminator is mutually influenced, and their optimization goals are competing with each other.

4. Generated image quality: Finally, the image sample generated by the sample_images function can be used to evaluate the effect of the generator training. If the generated image quality is high and similar to the real image, it means that the generator training effect is good.

### 2.6 Generation distribution map

By comparing the distribution of the generated image with the real sample, the training effect of the generator is evaluated. Using the sns.distplot function, the distribution curves of pixel values of the real image and the generated image are plotted respectively, and then compared and displayed. If the two distributions overlap more, it means that the generated image has a higher distribution similarity to the real image.

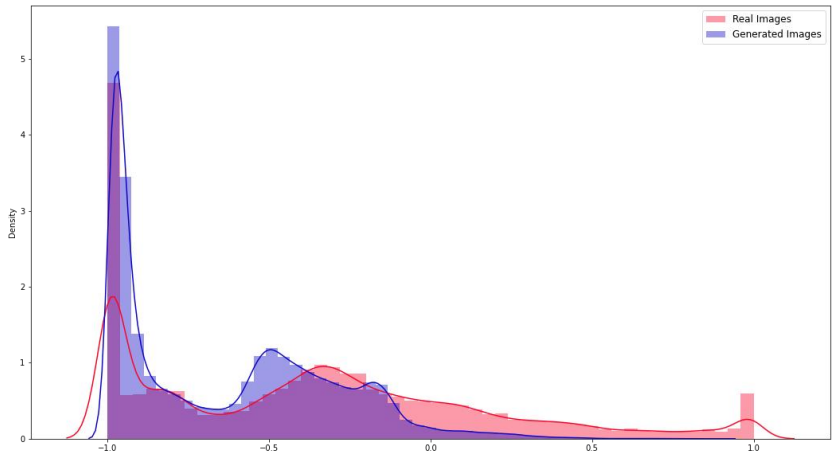

**figure 4.**Generate a distribution map of the sample

Combining generative intelligence with medical image processing technology, we can draw the following experimental conclusions: By using generative models, we can achieve the enhancement, reconstruction and synthesis of medical images, thereby improving the quality and quantity of images. [14-15]This technology can be used to improve medical imaging diagnostics, personalized medicine, and surgical simulation and planning. The distribution similarity between the generated image and the real image is high, indicating that the generated image can truly reflect the patient's condition, provide doctors with more comprehensive diagnostic information, promote the development of precision medicine, and improve the level of medical service and the treatment experience of patients. Therefore, the application of the combination of generative intelligence and medical image processing technology will bring significant improvements and sustainable benefits to the real-world medical field.

## 3    CONCLUSION

The integration of generative intelligence with medical image processing technology presents a transformative opportunity in the medical field. Through the synthesis, enhancement, and reconstruction of medical images, this technology enhances both the quality and quantity of available diagnostic data. The high similarity between generated and real images reflects the patient's condition accurately, offering healthcare professionals comprehensive diagnostic insights. This advancement promotes precision medicine, improves medical service quality, and enhances patient treatment experiences, promising significant and sustainable benefits in real-world medical practice.Experimental results demonstrate the effectiveness of generative models in augmenting medical image datasets and improving the quality of generated images. By leveraging techniques like Generative Adversarial Networks (GANs), these models address challenges associated with limited and heterogeneous datasets in medical imaging. Through image synthesis, enhancement, and reconstruction, generative AI facilitates more accurate diagnoses, personalized medicine approaches, and enhanced surgical planning. The high distribution similarity between generated and real images underscores the potential of this technology to advance medical diagnostics and patient care, heralding a new era in healthcare innovation.

Medical artificial intelligence such as Al medical imaging has entered a change, and the product variety has increased, but the research and development investment and income do not necessarily match. Product life cycle management has become the key to the success or failure of enterprises, including scientific research foundation, clinical evaluation, commercial landing and ecological pattern. Enterprises should rationally, prudently and comprehensively evaluate the future market space of products to avoid wasting time and money. Comb seven ecological routes of Al medical imaging enterprises. Namely: (1) Construct surgical robot + artificial intelligence medical image ecological route; (2) Construct medical information + artificial intelligence medical image ecological route; (3) Construct an integrated diagnosis and treatment + artificial intelligence medical image ecological route; (4) Construct an ecological route for AI medical images to go to sea; (5) Construct an ecological route for AI medical imaging products to enter medical insurance; (6) Construct the consumer side artificial intelligence medical image ecological route; (7) Construct portable devices + artificial intelligence medical imaging ecological route

In the future trend, generative artificial intelligence will bring exponential growth to Al medical imaging enterprises, and the combination of comprehensive medical artificial intelligence models and medical imaging fields will release great potential. Therefore, AI medical imaging enterprises will rely on the ecological route to accelerate their own hematopoietic speed and improve the ability to commercialize.